\documentclass[a4paper,twoside]{article}

\usepackage{epsfig}
\usepackage{subcaption}
\usepackage{calc}
\usepackage{amssymb}
\usepackage{amstext}
\usepackage{amsmath}
\usepackage{amsthm}
\usepackage{multicol}
\usepackage{pslatex}
\usepackage{apalike}
\usepackage{algorithm2e}
\usepackage{graphicx}
\usepackage[bottom]{footmisc}
\usepackage{tabularx}
\usepackage{array}
\usepackage{multirow}
\usepackage{booktabs}
\usepackage[colorlinks=false, linkcolor=black, citecolor=black, urlcolor=black]{hyperref}
\usepackage{geometry}
\usepackage[dvipsnames]{xcolor}
\usepackage{tikz}
\usepackage[edges]{forest}
\usepackage{SCITEPRESS}     

\begin{document}

\title{Bridging the Dimensionality Gap: A Taxonomy and Survey of 2D Vision Model Adaptation for 3D Analysis}

\author{\authorname{Akshat Pandya\sup{1}\orcidAuthor{0009-0008-8509-1760}, Bhavuk Jain\sup{1}\orcidAuthor{0009-0001-4304-2468}}
\affiliation{\sup{1} Independent Researcher}
\affiliation{\{akshatpandya97, jainbhavuk630\}@gmail.com}
\affiliation{{Equal contribution, names in alphabetical order}}
}

\keywords{3D Computer Vision, 3D Deep Learning, 2D to 3D Adaptation, Deep Learning, Convolutional Neural Networks, Vision Transformers, Survey, Taxonomy.}

\abstract{The remarkable success of Convolutional Neural Networks (CNNs) and Vision Transformers (ViTs) in 2D vision has spurred significant research in extending these architectures to the complex domain of 3D analysis. Yet, a core challenge arises from a fundamental dichotomy between the regular, dense grids of 2D images and the irregular, sparse nature of 3D data such as point clouds and meshes. This survey provides a comprehensive review and a unified taxonomy of adaptation strategies that bridge this gap, classifying them into three families: (1) Data-centric methods that project 3D data into 2D formats to leverage off-the-shelf 2D models, (2) Architecture-centric methods that design intrinsic 3D networks, and (3) Hybrid methods, which synergistically combine the two modeling paradigms to benefit from both rich visual priors of large 2D datasets and explicit geometric reasoning of 3D models. Through this framework, we qualitatively analyze the fundamental trade-offs between these families concerning computational complexity, reliance on large-scale pre-training, and the preservation of geometric inductive biases. We discuss key open challenges and outline promising future research directions, including the development of 3D foundation models, advancements in self-supervised learning (SSL) for geometric data, and the deeper integration of multi-modal signals.}

\onecolumn \maketitle \normalsize \setcounter{footnote}{0} \vfill

\section{\uppercase{Introduction}}
\label{sec:introduction}

\noindent The last decade witnessed a deep learning revolution in 2D computer vision. The success of Convolutional Neural Networks (CNNs) \cite{Krizhevsky12} and Vision Transformers (ViTs) \cite{Dosovitskiy20}, fueled by massive datasets like ImageNet \cite{Deng09} and LAION \cite{Schuhmann22}, established a paradigm of learning powerful visual features directly from data.

Given this success in 2D, a central question is how to best leverage these advancements for 3D analysis. The challenge is shaped by competing motivations. A pragmatic approach favors adapting proven 2D solutions to 3D problems to reuse the mature ecosystem of models and tools. However, this often conflicts with the need to preserve geometric fidelity, which is critical in applications like medical imaging where projecting 3D data into 2D views can cause unacceptable information loss. A third motivation seeks a synergistic fusion of both domains, injecting the rich semantic understanding from pre-trained 2D models into systems that can reason about detailed 3D geometry, a valuable approach given the relative lack of 3D data.

The core of this challenge lies in a fundamental mismatch between the structure of 2D data and that of 3D data—a problem we term the ``dimensionality gap". 2D architectures are designed for the dense, regular, and ordered grid structure of an image. In contrast, native 3D representations, such as point clouds and meshes, are inherently unstructured. A point cloud is a sparse, irregular, and unordered set of coordinates, requiring any processing model to be permutation invariant \cite{Qi17}. Similarly, a 3D mesh is topologically irregular. A standard 2D CNN or ViT cannot be directly applied to these raw 3D data because the foundational assumptions of a fixed neighborhood and an ordered input sequence are absent. Bridging this gap has been the primary catalyst for the diverse adaptation strategies that form the basis of this survey.

Existing 3D vision surveys typically organize the literature by data representation \cite{Ioannidou17}, \cite{Guo20}, task \cite{Arnold19}, \cite{Zhao24bev}, or architecture \cite{Lu22} \cite{Bronstein21}. Surveys on multi-modal fusion \cite{Feng20} \cite{Alaba22} are closely related but do not cover adaptation strategies that work without explicit multi-sensor data, such as multi-view rendering. This survey distinguishes itself by centering its analysis on the "dimensionality gap". We introduce a novel taxonomy—data-centric, architecture-centric, and hybrid—that organizes the field by the adaptation strategy itself, providing a unified framework to systematically compare the trade-offs between preserving geometric fidelity and leveraging powerful 2D pre-trained models.

To this end, this survey makes the following contributions:
\begin{enumerate}
    \item \textbf{A Novel Taxonomy:} We propose an intuitive taxonomy that organizes 2D-to-3D adaptation strategies into three families: Data-centric, Architecture-centric, and Hybrid methods.
    \item \textbf{A Structured and Systematic Review:} We provide a comprehensive review of seminal and contemporary works, using our taxonomy to categorize and explain the core principles behind different approaches.
    \item \textbf{A Critical Analysis of Trade-offs:} We analyze the fundamental trade-offs inherent to each category, focusing on compute, geometric fidelity, and the use of pre-trained 2D knowledge.
    \item \textbf{A Forward-Looking Perspective:} We identify key unresolved challenges and suggest promising avenues for future research, including 3D foundation models and deeper multi-modal fusion.
\end{enumerate}

In this survey, 3D analysis refers to interpreting existing spatial data (e.g., point clouds, meshes) for perception tasks by leveraging 2D data, architectures, or knowledge. We do not cover generative tasks like 3D shape synthesis. The paper is organized as follows. Section \ref{sec:preliminaries} covers the preliminaries, Section \ref{sec:taxonomy} presents our taxonomy, Section \ref{sec:analysis} analyzes the trade-offs, and Section \ref{sec:challenges} discusses future directions before we conclude in Section \ref{sec:conclusion}.

\section{\uppercase{Preliminaries}}
\label{sec:preliminaries}

\subsection{Pillars of 2D Vision}
Modern 2D computer vision relies primarily on two architectures- CNNs and ViTs. CNNs leverage local connectivity and weight sharing to efficiently learn hierarchical, translation-invariant features, making them highly effective for detecting local patterns like edges and textures. In contrast, ViTs process images as sequences of patches using self-attention mechanisms to model long-range dependencies and global context. While ViTs lack the inherent spatial inductive biases of CNNs and require large-scale pre-training, they offer superior scalability for holistic scene understanding. 

\subsection{The Landscape of 3D Data Representations}
Unlike 2D images, 3D data lacks a single canonical format. The choice of representation fundamentally dictates the adaptation strategy, balancing computational efficiency against geometric fidelity. The four primary categories are summarized in Table \ref{tab:3D_data_representations}.

\begin{table*}[t]
    \centering
    \caption{Comparison of 3D Data Representations}
    \label{tab:3D_data_representations}
    \begin{tabularx}{\linewidth}{@{} >{\centering\arraybackslash}p{0.13\linewidth} 
                                      >{\centering\arraybackslash}X 
                                      >{\centering\arraybackslash}X 
                                      >{\centering\arraybackslash}X @{}}
    \toprule
        \textbf{Representation} & \textbf{Description} & \textbf{Pros} & \textbf{Cons}\\
    \midrule
        Projective or View-based & Collection of 2D images from multiple viewpoints. & Allows direct reuse of mature 2D models; highly efficient. & Loses geometric information due to occlusion; not "true" 3D.\\
    \midrule
        Volumetric & 3D space discretized into a regular grid of voxels. & Regular structure enables straightforward 3D CNN extension. & Memory/compute scale cubically ($O(N^3)$); inefficient for sparse data.\\
    \midrule
        Geometric Primitives & Explicit surface geometry via Point Clouds (unordered coordinates) or Meshes (vertices/edges). & Memory-efficient; captures explicit surface topology. & Irregular structure violates standard CNN assumptions.\\
    \midrule
        Implicit & Continuous functions (e.g., NeRFs \cite{Mildenhall21}, SDFs \cite{Park19}) parameterized by networks. & Resolution-free; compact encoding of complex shapes. & Slow query speeds; better for synthesis than real-time analysis.\\
    \bottomrule
    \end{tabularx}
\end{table*}

\section{\uppercase{A Taxonomy of 2D-to-3D Adaptation Strategies}}
\label{sec:taxonomy}

\begin{figure*}[t]
  \centering
   \includegraphics[width=10cm]{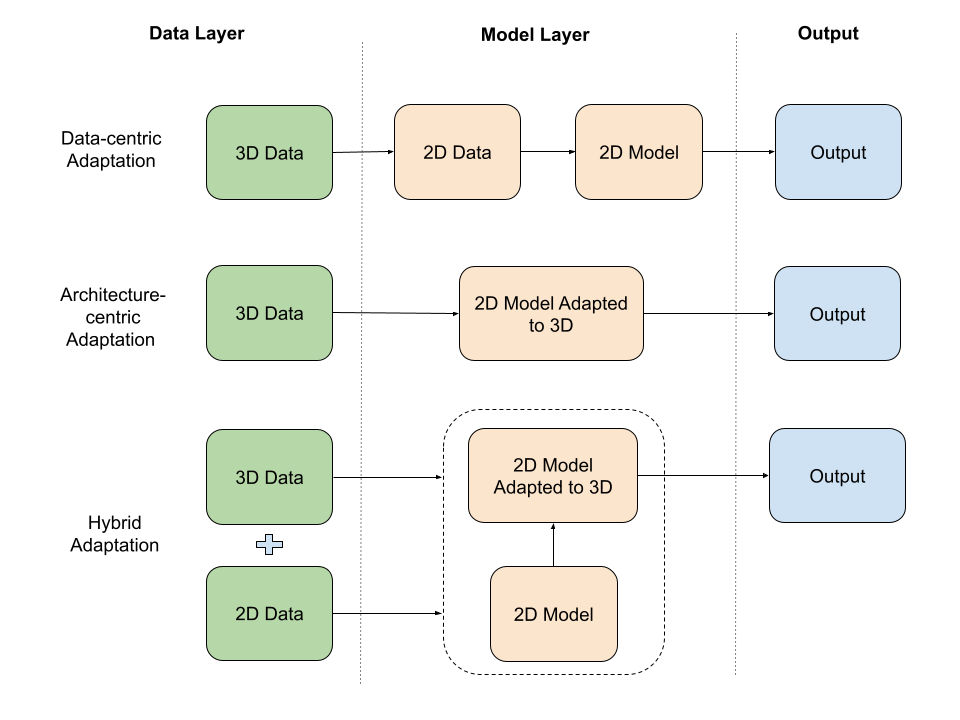}
   \caption{Illustration of the taxonomy of 2D-to-3D adaptation strategies.}
  \label{fig:taxonomy_overview}
\end{figure*}

\noindent This section presents a taxonomy for adapting 2D vision models to 3D, categorizing methods by their strategies for bridging the dimensionality gap. As shown in Figure \ref{fig:taxonomy_overview}, we have categorized the methods into three categories- data-centric, architecture-centric and hybrid adaptation. This unified framework provides a structured lens for analyzing state-of-the-art methods, emphasizing their core principles, strengths, and limitations. The taxonomy applies broadly across various fields in computer vision but focuses on common tasks such as object recognition, segmentation, and detection to illustrate key ideas, rather than exhaustively covering all possible applications.

\subsection{Data-centric Adaptation}
\label{subsec:data_centric}

The most straightforward approach to bridge the dimensionality gap is to directly adopt successful 2D model architectures. Data-centric adaptation methods transform 3D data into 2D or 2D-like formats. The core idea is to reproject, reorganize, or otherwise represent sparse and irregular 3D data as dense or sparse regular grids. This enables the direct use of 2D paradigms and pre-trained 2D CNN and transformer models. These methods are often simple and efficient, providing strong performance by inheriting rich features learned from massive 2D datasets.

\subsubsection{Multi-View Rendering}

Multi-view rendering represents the most established and intuitive data-centric approach. This approach was pioneered by the Multi-View Convolutional Neural Network (MVCNN) \cite{Su15}, which introduced a streamlined pipeline that renders a 3D object from multiple viewpoints, processes each 2D image with a shared-weight CNN, and aggregates the resulting features using element-wise max pooling to produce a global shape descriptor. Subsequent research focused on developing more sophisticated aggregation strategies to incorporate richer spatial relationships. GVCNN \cite{Feng18} sought to enhance pooling effectiveness by grouping similar viewpoints to mitigate redundancy. View-GCN \cite{Wei20} further advanced this approach by abandoning simple pooling entirely and instead treating each view as a node in a graph, enabling the model to explicitly learn and leverage inter-view relationships for more sophisticated feature fusion. Recently, models like MVT \cite{Chen21mvt} have modernized this approach by replacing CNN backbones with ViTs, demonstrating the adaptability and continued relevance of the paradigm. 

\subsubsection{Voxelization and Projection}

This methodology, prevalent in autonomous driving applications, addresses the dimensionality challenge by discretizing 3D point cloud data into a grid format and projecting its features onto a 2D Bird's-Eye View (BEV) map, thereby facilitating the use of 2D CNNs. The evolution of this approach has been characterized by a trade-off between geometric fidelity and computational efficiency. The pioneering method, VoxelNet \cite{Zhou18}, segmented the point cloud into 3D voxels, which were then processed using a 3D CNN, establishing the fundamental pipeline, albeit at a considerable computational expense. Subsequent research efforts have focused on optimization, with SECOND \cite{Yan18} introducing sparse 3D convolutions to enhance efficiency by processing only non-empty voxels. PointPillars \cite{Lang19}, on the other hand, entirely eliminates 3D convolutions and instead groups points into vertical pillars to directly generate a 2D pseudo-image, demonstrating a highly effective and faster alternative. Recent advancements, as exemplified by BEVFormer \cite{Li24}, have moved to incorporating multiple sensor modalities using Transformers to dynamically query and integrate features from multiple camera views into a unified and semantically rich BEV representation.

\subsubsection{Geometric Unfolding and Spherical Projection}

This sub-category of methods creates a 2D "pseudo-image" by mapping 3D data onto a grid using a projection that aims to preserve intrinsic geometric structure, in contrast to the simple top-down projection of BEV methods. The literature demonstrates two distinct parallel strategies for this mapping, often dictated by the underlying 3D data representation and the target application. For analyzing 3D meshes where surface topology is critical, methods like Geodesic CNNs \cite{Masci15} create local 2D patches by "unfolding" the mesh surface using geodesic distance as a metric, thus preserving the true local structure of the object's shape. Alternatively, for applications involving raw sensor data, particularly from LiDAR, a sensor-centric spherical projection is common. For example, RangeNet++ \cite{Milioto19} directly leverages the spherical nature of a LiDAR scan by projecting the point cloud onto a 2D range image, where pixel coordinates correspond to azimuth and elevation angles and pixel values represent distance.

\subsection{Architecture-centric Adaptation}
\label{subsec:arch_centric}

In contrast to data-centric approaches, Architecture-centric methods tackle the dimensionality gap by designing novel network architectures or modifying existing ones to inherently process raw 3D data. This approach often leads to better preservation of fine-grained geometric detail and can learn more complex 3D inductive biases. However, it typically forgoes the direct benefits of large-scale 2D pre-training.

\definecolor{boxblue}{HTML}{D9E8F7}
\definecolor{boxyellow}{HTML}{FBF0D5}
\definecolor{boxpurple}{HTML}{E9D8F2}
\definecolor{boxgrey}{HTML}{F0F0F0}
\tikzset{
    base/.style = {rectangle, rounded corners, draw=black, text width=1.75cm, minimum height=1cm, text centered, align=center, font=\fontsize{4}{4}\selectfont, inner sep=0pt, outer sep=0pt},
    main_branch/.style = {base, fill=boxgrey},
    box_blue/.style = {base, fill=boxblue, text width=1.75cm},
    box_yellow/.style = {base, fill=boxyellow, text width=1.75cm},
    box_purple/.style = {base, fill=boxpurple, text width=1.75cm},
    title_box/.style = {base, minimum height=2cm, text width=6.5cm},
    item_box/.style = {base, text width=4.5cm, inner sep=2pt},
    arrow/.style = { ->, >=stealth},
    line/.style = {thick}
}

\begin{figure*}[h!]
\centering
\resizebox{\textwidth}{!}{
\begin{tikzpicture}[node distance=1cm]

\node (root) [main_branch] {2D adaptations for 3D analysis};

\coordinate[right of=root, xshift=-0.05cm] (branch_point_arrow);
\coordinate[right of=root, xshift=1cm] (branch_point);

\node (data)      [main_branch, above of=branch_point, yshift=2.9cm] {Data-centric Adaptation};
\node (model)     [main_branch, at=(branch_point)] {Architecture-centric Adaptation};
\node (hybrid)    [main_branch, below of=branch_point, yshift=-4.7cm] {Hybrid Adaptation};

\coordinate[right of=data, xshift=-0.05cm] (branch_point_data_arrow);
\coordinate[right of=model, xshift=-0.05cm] (branch_point_model_arrow);
\coordinate[right of=hybrid, xshift=-0.05cm] (branch_point_hybrid_arrow);

\node (multi-view) [box_blue, right of=data, xshift=1cm, yshift=1.23cm] {Multi-view Rendering};
\node (voxel)      [box_blue, right of=data, xshift=1cm] {Voxelization and Projection};
\node (unfolding) [box_blue, right of=data, xshift=1cm, yshift=-1.35cm] {Geometric Unfolding and Spherical Projection};

\node (conv3d)    [box_yellow, right of=model, xshift=1cm, below of=model, yshift=2.38cm] {3D Convolutions on Voxel Grids};
\node (point-net) [box_yellow, right of=model, xshift=1cm, below of=model, yshift=1.35cm] {Point-based Networks};
\node (transform) [box_yellow, right of=model, xshift=1cm, below of=model, yshift=0.26cm] {3D Transformers};
\node (graph-net) [box_yellow, right of=model, xshift=1cm, below of=model, yshift=-0.83cm] {Graph-based Networks};

\node (guided)  [box_purple, right of=hybrid, xshift=1cm, yshift=2.41cm] {2D-Guided 3D Feature Learning};
\node (distill) [box_purple, right of=hybrid, xshift=1cm] {Cross-Modal Knowledge Distillation};
\node (fusion)  [box_purple, right of=hybrid, xshift=1cm, yshift=-2.26cm] {Multi-Modal Fusion Architectures};

\node (mv-title) [item_box, fill=boxblue, right of=multi-view, xshift=2.5cm] {
MVCNN \cite{Su15} \\
GVCNN \cite{Feng18} \\
RotationNet \cite{Kanezaki18} \\
View-GCN \cite{Wei20} \\
PointCLIP \cite{Zhang22pointclip} \\
MVT \cite{Chen21mvt}};

\node (voxel-title) [item_box, fill=boxblue, right of=voxel, xshift=2.5cm] {
VoxelNet \cite{Zhou18} \\
SECOND \cite{Yan18} \\
PointPillars \cite{Lang19} \\
PolarNet \cite{Zhang20} \\
BEVFormer \cite{Li24}\\
Pixor \cite{Yang18pixor} \\
2DPASS \cite{Yan22} \\
Fusing Bird View LIDAR Point Cloud and Front View Camera Image for Deep Object Detection \cite{Wang17fusing}
};

\node (unfolding-title) [item_box, fill=boxblue, right of=unfolding, xshift=2.5cm] {
Geodesic CNNs \cite{Masci15} \\
RangeNet++ \cite{Milioto19} \\
Spherical CNNs \cite{Cohen18} \\
Point2Pix \cite{Hu23} \\
Deep Projective 3D Semantic Segmentation \cite{Lawin17} \\
SqueezeSeg \cite{Wu18} \\
Pix4Point \cite{Qian22} \\
2DPASS \cite{Yan22} \\
};

\node (conv3d-title) [item_box, fill=boxyellow, right of=conv3d, xshift=2.5cm] {
3D ShapeNets \cite{Wu15}\\
VoxNet \cite{Maturana15}\\
3D U-Net \cite{Cicek16}\\
SegCloud \cite{Tchapmi17}\\
Minkowski Engine \cite{Choy19}\\
};

\node (point-net-title) [item_box, fill=boxyellow, right of=point-net, xshift=2.5cm] {
PointNet \cite{Qi17}\\
PointNet++ \cite{Qi17plus}\\
DGCNN \cite{Wang19}\\
KPConv \cite{Thomas19}\\
PointCNN \cite{Li18pointcnn}\\
Image2Point \cite{Xu22img2point}\\
};

\node (graph-net-title) [item_box, fill=boxyellow, right of=graph-net, xshift=2.5cm] {
MeshCNN \cite{Hanocka19}\\
GAT \cite{Velickovic17}\\
ClusterNet \cite{Chen19cluster}\\
Point-GNN \cite{Shi20}\\
View-GCN \cite{Wei20} \\
};

\node (transform-title) [item_box, fill=boxyellow, right of=transform, xshift=2.5cm] {
Point Transformer \cite{Zhao21}\\
PCT \cite{Guo21}\\
Point-BERT \cite{Yu22}\\
Stratified Transformer \cite{Lai22}\\
PVT \cite{She23}\\
Adapting Pre-trained Vision Transformers from 2D to 3D through Weight Inflation Improves Medical Image Segmentation \cite{Zhang22adapting}\\
};

\node (guided-title) [item_box, fill=boxpurple, right of=guided, xshift=2.5cm] {
PointPainting \cite{Vora20}\\
EPNet \cite{Huang20}\\
MVP \cite{Yin21}\\
3D-CVF \cite{Yoo20}\\
AutoAlign \cite{Chen22autoalign}\\
Image2Point \cite{Xu22img2point}\\
Adapt PointFormer \cite{Li24adapt}\\
Pix4Point \cite{Qian22}\\
Any2Point \cite{Tang24}\\
2DPASS \cite{Yan22}\\
PointCLIP \cite{Zhang22pointclip}\\
P2P \cite{Wang22p2p}\\
};

\node (distill-title) [item_box, fill=boxpurple, right of=distill, xshift=2.5cm] {
3D segmenter \cite{Wu22seg}\\
DINR \cite{Zeid25}\\
CLIP2Scene \cite{Chen23clip}\\
Distillbev \cite{Wang23distill}\\
Stxd \cite{Jang23}\\
Partdistill \cite{Umam24}\\
PointKAD \cite{Sanjay25}\\
I2P-MAE \cite{Zhang23learning}\\
ViT3D \cite{Sajid25}\\
ACT \cite{Dong22}\\
Learning from 2D: Contrastive Pixel-to-Point Knowledge Transfer for 3D Pretraining \cite{Liu21learning}\\
SliDR \cite{Sautier22}\\
CrossPoint \cite{Afham22}\\
SimIPU \cite{Li22simipu}\\
Joint-MAE \cite{Guo23}\\
Solving 3D Inverse Problems using Pre-trained 2D Diffusion Models \cite{Chung23}\\
2DPASS \cite{Yan22}\\
PiMAE \cite{Chen23pimae}\\
};

\node (fusion-title) [item_box, fill=boxpurple, right of=fusion, xshift=2.5cm] {
PointFusion \cite{Xu18}\\
MVX-Net \cite{Sindagi19}\\
BEVFusion \cite{Liu22bev}\\
DeepFusion \cite{Li22deep}\\
PiMAE \cite{Chen23pimae}\\
Fusing Bird View LIDAR Point Cloud and Front View Camera Image for Deep Object Detection \cite{Wang17fusing}\\
Adapt PointFormer \cite{Li24adapt}\\
Joint-MAE \cite{Guo23}\\
SAM 3D \cite{chen2025sam}\\
};
\draw [line] (root) -- (branch_point_arrow.west);
\draw [arrow] (branch_point_arrow.east) |- (data.west);
\draw [arrow] (branch_point_arrow.east) -- (model.west);
\draw [arrow] (branch_point_arrow.east) |- (hybrid.west);

\draw [line] (data) -- (branch_point_data_arrow.west);
\draw [arrow] (branch_point_data_arrow.east) |- (multi-view.west);
\draw [arrow] (branch_point_data_arrow.east) |- (voxel.west);
\draw [arrow] (branch_point_data_arrow.east) |- (unfolding.west);

\draw [line] (model.east) -- (branch_point_model_arrow.west);
\draw [arrow] (branch_point_model_arrow.east) |- (conv3d.west);
\draw [arrow] (branch_point_model_arrow.east) |- (point-net.west);
\draw [arrow] (branch_point_model_arrow.east) |- (graph-net.west);
\draw [arrow] (branch_point_model_arrow.east) |- (transform.west);

\draw [line] (hybrid.east) -- (branch_point_hybrid_arrow.west);
\draw [arrow] (branch_point_hybrid_arrow.east) |- (guided.west);
\draw [arrow] (branch_point_hybrid_arrow.east) |- (distill.west);
\draw [arrow] (branch_point_hybrid_arrow.east) |- (fusion.west);

\draw [arrow] (multi-view.east) -- (mv-title.west);
\draw [arrow] (voxel.east) -- (voxel-title.west);
\draw [arrow] (unfolding.east) -- (unfolding-title.west);

\draw [arrow] (conv3d.east) -- (conv3d-title.west);
\draw [arrow] (point-net.east) -- (point-net-title.west);
\draw [arrow] (transform.east) -- (transform-title.west);
\draw [arrow] (graph-net.east) -- (graph-net-title.west);

\draw [arrow] (guided.east) -- (guided-title.west);
\draw [arrow] (distill.east) -- (distill-title.west);
\draw [arrow] (fusion.east) -- (fusion-title.west);

\end{tikzpicture}
}

\caption{A flowchart representing our taxonomy of 2D-to-3D adaptation methods.}
\label{fig:flowchart_tikz}
\end{figure*}
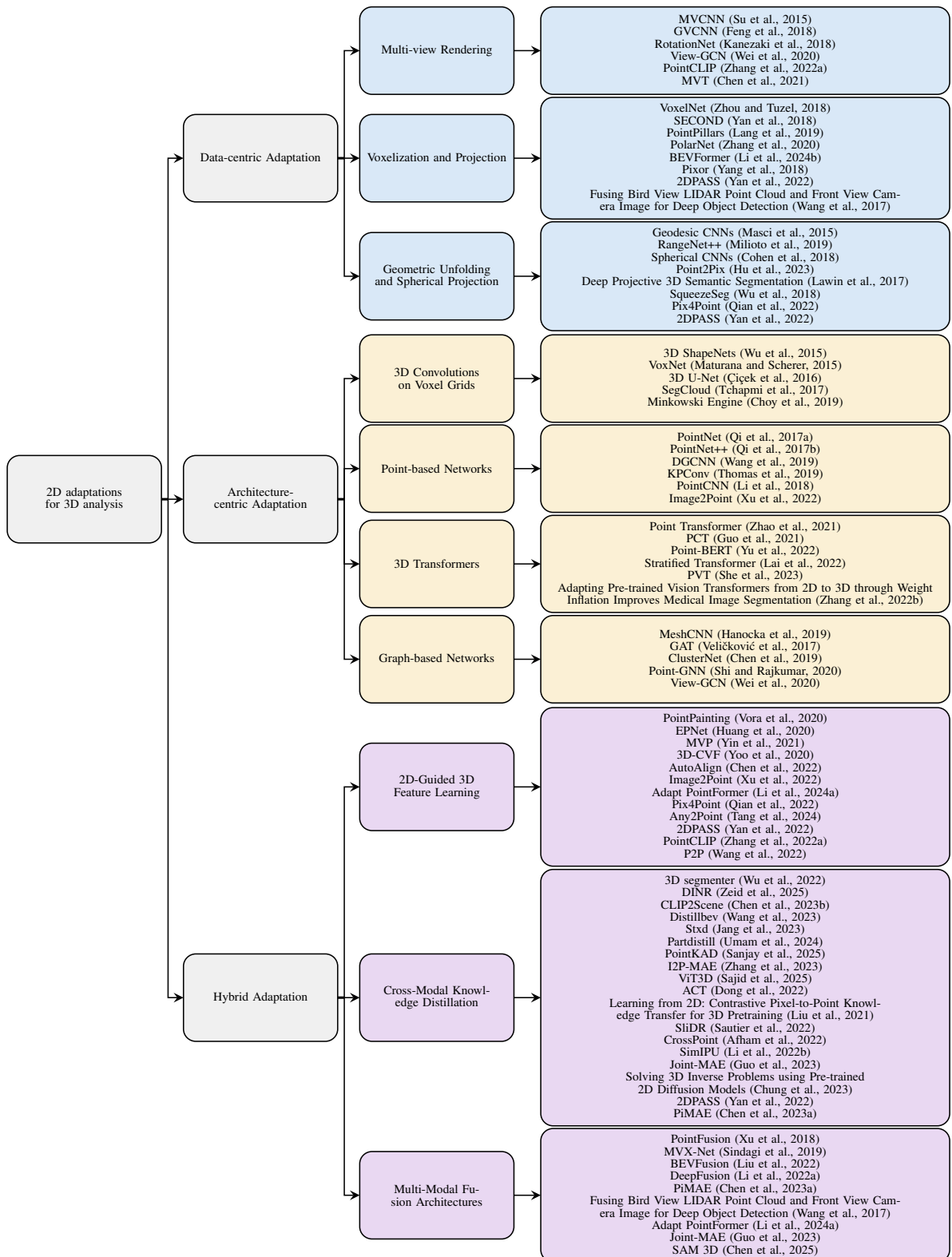

\subsubsection{3D Convolutions on Voxel Grids}

The most straightforward architectural extension of 2D CNNs to 3D involves the application of 3D convolutions on a volumetric voxel grid. Pioneering works, such as 3D ShapeNets \cite{Wu15} and VoxNet \cite{Maturana15}, initially demonstrated the feasibility of this approach for 3D object recognition. However, these methods were fundamentally limited by the cubic growth ($O(N^3)$) in memory and computation, restricting their application to low-resolution grids, which became a central challenge in the field. The critical advancement came with the development of submanifold sparse convolutions \cite{Graham17}, an innovation popularized by libraries like the Minkowski Engine \cite{Choy19}. This approach recognized the inherent sparsity of most 3D data and redefined the convolution operation to target only non-empty voxels, thereby addressing the computational bottleneck of dense convolutions and enabling the development of significantly deeper and higher-resolution models. This made influential architectures, such as the 3D U-Net \cite{Cicek16}, practical for high-resolution volumetric medical image segmentation, unlocking the use of 3D CNNs for a broad array of tasks.

\subsubsection{Point-based Networks}

This class of architectures operates directly on raw point clouds, avoiding any explicit conversion to a grid-based format. The key challenge in this domain is permutation invariance, arising from the unordered nature of point clouds. The seminal PointNet \cite{Qi17} provided an initial solution by processing each point independently through a shared network, followed by a symmetric aggregation function, such as max pooling, to generate a global feature vector. However, a critical limitation of this approach was its inability to capture fine-grained local geometric structure, as it did not explicitly model relationships between neighboring points. Subsequent research focused on addressing this specific limitation. PointNet++ \cite{Qi17plus} introduced a hierarchical architecture that recursively applies the PointNet operator on nested partitions of the point set, enabling the learning of local features across multiple scales. Other works have proposed distinct strategies for the same purpose; for example, DGCNN \cite{Wang19} dynamically constructs graphs in feature space at each layer to capture more meaningful semantic relationships between points, going beyond purely spatial correlations. In a complementary effort to more closely emulate traditional convolutions, KPConv \cite{Thomas19} designed a flexible and deformable convolution operator for point clouds, defined by a set of learnable kernel points that adapt to the underlying geometry of the data.

\subsubsection{Graph-based Networks}

These methods process 3D data by representing it as a graph $G = (V, E)$ and applying a Graph Neural Network (GNN) to learn features via message passing between connected nodes. The literature reveals two primary parallel strategies based on whether the input data has an explicit or implicit graph structure. For 3D meshes, which have a predefined topology, methods like MeshCNN \cite{Hanocka19} leverage this fixed structure by defining specialized convolutions that operate directly on the mesh's edges, effectively exploiting the explicit surface connectivity for tasks like shape segmentation. In contrast, for unstructured point clouds, the graph must first be constructed. A powerful approach is demonstrated by DGCNN \cite{Wang19}, which goes beyond simple spatial connections by dynamically recomputing the graph in feature space at each layer of the network, allowing point relationships to be defined by learned semantic similarity rather than just proximity. A key architectural innovation applicable to both strategies is the use of attention, as pioneered by the Graph Attention Network (GAT) \cite{Velickovic17}, which enables a node to weigh the importance of messages from its neighbors, leading to more expressive feature aggregation. 

\subsubsection{3D Transformers}

This class of architectures adapts the attention mechanism of the ViT \cite{Dosovitskiy20} to operate directly on 3D point sets. Early works, such as the Point Transformer \cite{Zhao21}, demonstrated the power of applying self-attention to local neighborhoods of points, effectively capturing contextual geometric relationships. However, the standard transformer's major limitation is its quadratic computational complexity ($O(N^2)$) with respect to the number of input tokens, which makes applying global attention to large-scale point clouds computationally prohibitive. Subsequent research has progressed along two primary fronts to address this core challenge. The first focuses on architectural efficiency, with models like the Stratified Transformer \cite{Lai22} introducing novel attention mechanisms that operate on a sparse, downsampled set of key points to achieve scalability while retaining long-range dependencies. The second, parallel direction focuses on data efficiency through self-supervised pre-training. Point-BERT \cite{Yu22}, for example, successfully adapted the masked modeling paradigm from natural language processing, training the model to reconstruct masked portions of a point cloud. This allows the network to learn robust geometric priors from massive unlabeled datasets, significantly improving performance on downstream tasks. 

\subsection{Hybrid Adaptation}
\label{subsec:hybrid}

Hybrid methods represent a compelling synthesis of the two preceding categories. They synergistically combine the pre-trained feature representations from 2D vision models with the geometric specificity of intrinsic 3D architectures.

\subsubsection{2D-Guided 3D Feature Learning}

This paradigm enriches raw 3D point data by augmenting it with semantic features extracted from corresponding 2D images. PointPainting \cite{Vora20} established this idea by running a 2D semantic segmentation network on images and appending the predicted class scores to the aligned LiDAR points. Later works developed more integrated fusion strategies. EPNet \cite{Huang20}, for example, enhances point features with multi-scale image features through tightly coupled modules. However, such methods remain sensitive to calibration errors between sensors. AutoAlign \cite{Chen22autoalign} mitigates this by learning an alignment map to automatically align multi-modal features. A parallel strategy instead focuses on directly adapting the 2D model by fine-tuning the pre-trained 2D architecture for 3D analysis. Image2Point \cite{Xu22img2point} inflates a pre-trained 2D network into 3D and fine-tunes it for point cloud tasks, while methods like P2P \cite{Wang22p2p} and Adapt PointFormer \cite{Li24adapt} repurpose pre-trained ViTs by introducing lightweight prompts or adaptation modules that enable direct 3D understanding. 

\subsubsection{Cross-Modal Knowledge Distillation}

In this paradigm, a 3D "student" network mimics a "teacher" model operating on a different, more informative modality. Two primary strategies exist for this cross-modal transfer, defined by the teacher model. The first, seen in DistillBEV \cite{Wang23distill}, uses an accurate LiDAR-based teacher to train a camera-only 3D student, enabling the final deployed system to achieve low latency using only cameras. The second, increasingly popular strategy distills knowledge from 2D pre-trained foundation models. For instance, DITR \cite{Zeid25} and CLIP2Scene \cite{Chen23clip} leverage the rich semantic understanding of DINOv2 \cite{oquab2023dinov2} and CLIP \cite{radford2021learning}, respectively, to supervise the 3D student's feature space. This injects rich, open-world semantic knowledge from the 2D domain into the 3D network, overcoming the scarcity of labeled 3D data.

\subsubsection{Multi-Modal Fusion Architectures}
Unlike 2D-guided methods, multimodal fusion architectures use parallel backbones for 2D and 3D data, with explicit modules to simultaneously integrate feature streams at various depths. Early works like PointFusion \cite{Xu18} exemplified a late fusion approach, where the point cloud and image were processed through separate deep encoders, and the final global feature vectors were combined before the prediction head. The limitation of this strategy is that the fusion occurs too late to allow for the learning of rich, co-dependent features. Subsequent research shifted toward intermediate fusion strategies, like BEVFusion \cite{Liu22bev}, which projects multi-modal features into a unified BEV representation mid-network, allowing for a tighter and more spatially-aware integration. The most recent evolution is toward deep fusion using attention mechanisms, as seen in DeepFusion \cite{Li22deep}. This approach utilizes cross-attention to allow features from one modality to dynamically query and integrate information from the other at multiple network layers, enabling a more profound fusion.

\section{\uppercase{Comparative Analysis and Discussion}}
\label{sec:analysis}

\noindent The taxonomic framework presented in this survey allows for a critical analysis of the fundamental trade-offs that researchers and practitioners must navigate. The choice between data-centric, architecture-centric, and hybrid methods is not merely one of preference but is dictated by a set of competing priorities. We dissect these key tensions below.

\subsection{Performance vs. Efficiency}

A primary trade-off in 3D deep learning lies between geometric accuracy and computational efficiency. Architecture-centric methods often achieve high geometric fidelity, but at a substantial cost. Early 3D convolutional models \cite{Maturana15} were constrained to low-resolution voxel grids, and while sparse convolutions \cite{Choy19} improved efficiency, high-resolution 3D operations remain expensive. Point-based methods like KPConv \cite{Thomas19} and 3D Transformers such as Point-BERT \cite{Yu22} also reach state-of-the-art performance but suffer from costly neighborhood searches or quadratic attention complexity. In contrast, data-centric methods favor efficiency by converting 3D problems to 2D. PointPillars \cite{Lang19} trades geometric detail for fast 2D CNN inference, crucial in autonomous driving, while multi-view methods like MVCNN \cite{Su15} achieve similar gains but introduce rendering overhead. Hybrid methods strike a middle ground, combining 2D and 3D streams for performance boosts at the expense of higher resource demands (e.g., BEVFusion \cite{Liu22bev}).

\subsection{Geometric Inductive Bias vs. Pre-training}

A second key tension lies between embedding strong geometric priors in the model architecture and leveraging semantic priors from large-scale 2D pre-training. Architecture-centric models, such as PointNet \cite{Qi17} and PointNet++ \cite{Qi17plus}, have innate geometric biases about the structure of point clouds, while MeshCNN \cite{Hanocka19} has a built-in bias for edge-based operations for meshes. These priors enable effective geometric analysis but require learning semantic concepts from comparatively smaller 3D datasets. In contrast, data-centric and hybrid approaches sacrifice geometric purity to harness semantic knowledge from 2D models. For example, PointPainting \cite{Vora20} uses semantic labels from a pre-trained 2D network. Hybrid methods, like cross-modal distillation in CLIP2Scene \cite{Chen23clip}, attempt to transfer knowledge from 2D foundation models to 3D networks, acknowledging the need for both geometric and semantic knowledge, with how best to merge them remaining a key research question.

\subsection{Application Suitability}
Ultimately, the choice of methodology is heavily dictated by the specific application. There is no universally superior approach; rather, each category excels under different constraints and task requirements.
\begin{enumerate}
    \item \textbf{Fine-grained Shape Analysis:} For tasks requiring a detailed understanding of an object's surface, such as part segmentation, correspondence analysis, or shape retrieval, architecture-centric methods are dominant. The high geometric fidelity of point-based networks like PointNet++ \cite{Qi17plus} and graph-based models like MeshCNN \cite{Hanocka19} is essential for reasoning about local and global shape properties.
    \item \textbf{Autonomous Driving and Robotics:} This domain is dominated by hybrid and efficient data-centric (BEV) methods. The task demands real time processing of large-scale, multi-modal sensor data. While BEV models like PointPillars \cite{Lang19} offer strong efficiency baselines, state-of-the-art performance now comes from hybrid fusion architectures such as BEVFusion \cite{Liu22bev} and DeepFusion \cite{Li22deep}, which show that robust perception in complex environments requires combining 2D semantic context with 3D spatial accuracy.    
    \item \textbf{Object Classification:} For the simpler task of classifying isolated and complete 3D shapes (e.g., on ModelNet/ShapeNet benchmarks), data-centric (multi-view) methods remain highly competitive. Approaches like MVCNN \cite{Su15} are often sufficient to capture the discriminative global characteristics of a shape and are relatively simple to implement.
    \item \textbf{Medical Image Analysis:} This field overwhelmingly favors architecture-centric (3D convolutions). The input data from CT or MRI scans is inherently volumetric, and preserving the full 3D spatial context is non-negotiable for tasks like tumor segmentation. The 3D U-Net \cite{Cicek16} architecture and its variants have become a canonical standard for this reason.
\end{enumerate}

\section{\uppercase{Open Challenges and Future Directions}}
\label{sec:challenges}

\noindent While significant progress has been made in bridging the dimensionality gap, the field is rapidly evolving, with several fundamental challenges and exciting research frontiers emerging. Based on our analysis, we identify the following promising directions that will likely shape the future of 3D computer vision.

\subsection{Quest for 3D Foundation Models}
A persistent challenge in 3D vision is the absence of an "ImageNet moment" — catalyzed by a massive, diverse dataset and pre-trained foundation models. Unlike in the 2D domain, acquiring and annotating large-scale, heterogeneous 3D data remains a significant bottleneck, and the lack of a dominant 3D representation complicates the design of a universal backbone. While Point-BERT \cite{Yu22} demonstrates the viability of pre-training for point clouds, future progress depends on curating massive, multi-representation 3D datasets and developing architectures that learn generalizable priors, potentially leveraging synthetic data.

\subsection{Advancements in Self-Supervised Learning}

Self-supervised learning is the most viable path for building 3D foundation models by reducing our dependence on limited labeled data. The field has advanced beyond basic tasks, adopting successful paradigms from other areas. Masked modeling, as seen in Point-BERT \cite{Yu22}, enables learning local geometry by reconstructing masked point cloud regions. Studies like STXD \cite{Jang23} utilize the teacher-student method for cross-modal distillation, and early methods like RotationNet \cite{Kanezaki18} demonstrated the value of geometric self-supervision. The future involves creating more advanced SSL objectives that teach models not only semantics, but also physics, affordances, and material properties from unlabeled 3D data.

\subsection{Deeper Multi-Modal Fusion}
While effective, many hybrid methods still use simple fusion strategies like decorating 3D points with 2D features (PointPainting \cite{Vora20}) or concatenating feature vectors (PointFusion \cite{Xu18}). The future lies in developing more deeply integrated fusion architectures, as seen with cross-attention mechanisms in DeepFusion \cite{Li22deep}. The scope of fusion is also expanding beyond RGB images and LiDAR to include natural language, as demonstrated by CLIP2Scene \cite{Chen23clip}, which paves the way for zero-shot 3D scene understanding.

\subsection{Large-Scale Dynamic Scenes}
Existing research primarily focuses on static, object-centric, or room-scale scenes. A major challenge is scaling these methods to city-scale and dynamic (4D) data. The computational complexity of intrinsic methods--specifically the $O(N^3)$ cost of dense 3D convolutions and the $O(N^2)$ cost of global self-attention--becomes a bottleneck. Architectures like the Stratified Transformer \cite{Lai22} are addressing this with more efficient attention mechanisms. Further, future models must explicitly reason about the temporal dimension. The temporal fusion in BEVFormer \cite{Li24} is a step in this direction. However, robustly modeling motion, tracking, and predicting future states in large-scale 4D point clouds is a vital research challenge, especially for applications like autonomous embodied systems and digital twins.

\subsection{Integration with Implicit Neural Representations}
Implicit Neural Representations, such as NeRFs \cite{Mildenhall21} and SDFs \cite{Park19}, have advanced 3D scene synthesis and reconstruction. However, a gap remains between these synthesis-oriented representations and feature-based architectures used for analysis and recognition. Implicit fields are computationally expensive to query and do not naturally produce the hierarchical feature maps that downstream recognition tasks rely on. A key future direction is bridging this gap, which could involve developing methods for recognition directly on implicit model weights, designing techniques to distill NeRFs into explicit representations, or creating unified architectures for both high-fidelity rendering and semantic understanding. This unification would be a major step toward a holistic 3D scene representation.

\section{\uppercase{Conclusion}}
\label{sec:conclusion}
\noindent This paper presents a comprehensive survey and a novel intellectual framework for adapting 2D vision models to 3D analysis. We began by establishing the motivation for this adaptation and identifying the "dimensionality gap" as the central challenge driving research. To provide a structured understanding of the field, we introduced a novel taxonomy that categorizes approaches into three distinct families: Data-centric, Architecture-centric, and Hybrid methods. Our analysis reveals that progress in this domain is governed by a persistent trade-off among computational efficiency, the preservation of geometric fidelity, and the effective use of semantic knowledge from large-scale 2D pre-training. Based on this review, we also identify several open challenges and future research directions. These include the pursuit of 3D foundation models via self-supervised learning, the development of more deeply integrated multi-modal fusion architectures, and the need to scale current methods to dynamic, large-scale environments.

\bibliographystyle{apalike}
{\small
\bibliography{example}}

\end{document}